\documentclass[10pt,journal,compsoc]{IEEEtran}

\usepackage{url}
\usepackage{multicol}
\usepackage{booktabs}       
\usepackage{amsfonts}       
\usepackage{nicefrac}       
\usepackage{microtype}      
\usepackage{wrapfig,verbatim}

\usepackage{varioref}

\usepackage[cmex10,tbtags]{amsmath}
\DeclareMathOperator*{\argmax}{argmax}
\DeclareMathOperator*{\argmin}{argmin}
\usepackage{cases}
\usepackage{amsfonts}
\usepackage{bbm}

\usepackage{paralist,xparse}
\usepackage{amsfonts}
\usepackage{multirow} 
\usepackage{subfigure}
\usepackage{amssymb} 
\usepackage{scrextend}
\usepackage{float}
\usepackage{adjustbox}

\usepackage{algorithm}
\usepackage{algorithmic}
\usepackage{mathtools}
\usepackage{times}
\usepackage{pgfplots}
\usepackage{array}

\usepackage{enumitem}
\usepackage{amsthm}

\definecolor{dgreen}{rgb}{0.0, 0.5, 0.0}

\newtheorem{theorem}{Theorem}
\newtheorem{definition}{Definition}

\newcommand{\CNN}{\text{CNN}}

\newcommand{\norm}[1]{\lVert#1\rVert} 
\newcommand{\stateset}{\ensuremath{\mathcal{X}}} 
\newcommand{\actionset}{\ensuremath{\mathcal{A}}} 
\newcommand{\prior}{\ensuremath{\pi_0}} 

\newcommand{\reals}{\ensuremath{\mathbb{R}}} 

\newcommand{\E}{\mathbb{E}}

\newcommand{\action}{a} 
\newcommand{\state}{x}

\newcommand{\obs}{y}

\newcommand{\obsset}{\mathcal{Y}}

\newcommand{\RIcost}{C}

\newcommand{\runcostinst}{c}

\newcommand{\utilityagent}[1]{\utilitysymbol_{#1}(\state,\action)}

\newcommand{\utilitysymbol}{u}

\newcommand{\utilitysymbolagent}[1]{\utilitysymbol_{#1}}

\newcommand{\eps}{\varepsilon}

\newcommand{\actiontwo}{b}

\newcommand{\datann}{\mathbb{D}}
\newcommand{\datasetaccum}{\datann}

\newcommand{\testimg}{s}
\newcommand{\testpred}{f}
\newcommand{\imgiter}{i}
\newcommand{\dpiter}{k}
\newcommand{\dpitertwo}{j}
\newcommand{\dpset}{\mathcal{K}}

\newcommand{\attfunsymb}{\alpha}
\newcommand{\attfunagent}[1]{\attfunsymb_{#1}}

\newcommand{\numdp}{K}

\newcommand{\robmet}{\mathcal{R}_{\text{\dtest}}}
\newcommand{\robmetcompact}{\mathcal{R}_{\text{\dtestcompact}}}

\newcommand{\uth}{^{\text{th}}}

\newcommand{\RImodel}{UMRI}
\newcommand{\RImodels}{S-UMRI}
\newcommand{\dtest}{BRP}
\newcommand{\dtestcompact}{S-BRP}

\newcommand{\bigtuple}{\Theta}

\newcommand{\numimg}{N}

\usepackage{mathtools}

\usepackage{tikz}
\usetikzlibrary{shapes.geometric, arrows}
\usepackage[european resistor, european voltage, european current]{circuitikz}
\usetikzlibrary{arrows,shapes,positioning,snakes,calc}
\usetikzlibrary{decorations.markings,decorations.pathmorphing,decorations.pathreplacing}
\usetikzlibrary{calc,patterns,shapes.geometric,topaths,fit}

\tikzset{
    block/.style={rectangle, draw, line width=0.5mm, black, text width=4.5em, text centered,
                 minimum height=1em},
               line/.style={draw, -latex}}

\title{Rationally Inattentive Utility Maximization for Interpretable Deep Image Classification}
\author{Kunal Pattanayak,~\IEEEmembership{Student Member,~IEEE}, Vikram Krishnamurthy,~\IEEEmembership{Fellow,~IEEE}\thanks{V. Krishnamurthy and K. Pattanayak are with the School
		of Electrical and Computer Engineering, Cornell University, Ithaca,
		NY, 14853 USA. e-mail: vikramk@cornell.edu, kp87@cornell.edu.}}

\pgfplotsset{compat=1.16}
\begin{document}
\maketitle

\begin{abstract} 
Are deep convolutional neural networks (CNNs)
for image classification explainable by utility
maximization with information acquisition costs?  
We demonstrate that deep CNNs behave equivalently
(in terms of necessary and sufficient conditions)
to rationally inattentive utility maximizers, a
generative model used extensively in economics
for human decision making.
Our claim is based by extensive experiments on 200 deep CNNs from 5 popular  architectures.
The parameters of our interpretable model are computed efficiently via convex feasibility algorithms.
As an application, we show that our economics-based interpretable model can predict the classification performance of deep CNNs trained with arbitrary parameters with accuracy exceeding $94\%$. This eliminates the need to re-train the deep CNNs for image classification.
The theoretical foundation of our approach lies in Bayesian revealed preference studied in micro-economics.
All our results are on GitHub and completely reproducible. 
\end{abstract}


\section{Introduction}
This paper considers interpretable deep image classification.\footnote{This paper builds substantially on our existing arXiv preprint https://arxiv.org/abs/2102.04594 uploaded in January, 2021.} We show that image classification using deep Convolutional Neural Networks (CNNs) can be interpreted as a generative human decision-making model developed in  microeconomics. 

In micro- and behavioral economics\footnote{Micro-economics models the interaction of individual agents pursuing their private interests.  Behavioral economics models human decision making in terms of subjective probabilities via prospect theory and framing. In the rest of this paper, we will use the term `agent' to denote a human decision-maker.}, a fundamental question relating to human decision making is: {\em How to model attention spans in humans (agents)?} The area of rational inattention~\cite{Sim03,Sim10}, pioneered by Nobel laureate Christopher Sims models human attention in information-theoretic terms. The key hypothesis is that agents  are ``boundedly rational"- their perception of the environment is modeled as a Shannon capacity limited channel. In simple terms, rational inattention assigns a mutual information cost for human attention spans.

Building on the rational inattention model, the next key concept is that of a Bayesian agent with rational inattention that maximizes its expected utility. Such models are studied extensively in~\cite{WD12,MAT15,DE17}. The intuition is this: more attentive decisions yield a higher expected utility at the expense of a larger attention cost. Hence, the Bayesian agent  optimally trades off between minimizing its sensing cost and maximizing its expected utility. An important question is: {\em How to test for rationally inattentive utility maximization given the decisions of a Bayesian agent?} In the last decade, necessary and sufficient conditions have been developed in the area of Bayesian revealed preference~\cite{CM15,CD15} to test if the decisions of a Bayesian agent are consistent with rationally inattentive utility maximization.
\subsection*{Summary of Results.} 
The question we address is: {\em Can the decisions of deep CNNs in image classification be explained by a rationally inattentive Bayesian utility maximizer?}


This paper uses a {\em data-driven} Bayesian revealed preference approach to interpretable deep CNN image classification.
Bayesian revealed preference performs a {\em post-hoc} analysis of agent decisions. It constructs a generative\footnote{A generative model is image-independent, and hence provides a global explanation for deep image classification. In contrast, local approximation models for deep image classification are image-specific; they approximate model decisions via tractable functionals in a $\delta$-neighborhood of every input.} explanatory model for the agent decisions, parameterized by
a utility function and an information acquisition cost. Our approach draws important parallels between human decision making and deep neural networks; namely that deep neural networks satisfy economics based rationality.

By very nature, Bayesian revealed preference reconstructs a {\em set} of feasible utility functions and information acquisition costs. Every element in the feasible set explains the deep CNN decisions equally well.
The computed utility function induces a preference ordering on the set of image classes. That is, how much a deep CNN prioritizes accurate classification over an inaccurate classification. The information acquisition cost abstracts the penalty incurred by the deep CNN to `learn' an accurate latent feature representation. The key results in this paper are:

\begin{figure}
\centering
\tikzstyle{block} = [rectangle, draw, fill=blue!20,     text width=12em, text centered, minimum height=4em]
\tikzstyle{nnblock} = [rectangle, draw, fill=red!25,     text width=2.5em, text centered, minimum height=2em, minimum width=2em]
\tikzstyle{bigblock} = [rectangle, draw, fill = yellow!20,    text width=16em, text centered, minimum height=7em]
\begin{tikzpicture}[auto, node distance=2cm,>=latex']
    \node [bigblock] (bigCNN) at (3.2,3.3) {};
    \node[align=center] at (3.2,4) {{\large{\bf Deep Convolutional}}\\ {\large{\bf Neural Network}}};
    
    \node [nnblock] (CNN) at (3,2.8) {{\bf CNN}};
    \node[name=ipimg,align=center] at (1.2,2.8) {\bf \textcolor{dgreen}{Input}\\ \bf \textcolor{dgreen}{Image} };
    \node[name=opimg,align=center] at (5,2.8) {\bf \textcolor{dgreen}{Predicted}\\ \bf \textcolor{dgreen}{Image} \\\bf \textcolor{dgreen}{Class} };
    \node [block, below of=bigCNN,
    node distance=3.4cm] (RI) {};
    \node[align=center] at (3.2,-0.1) {{\large {\bf Rationally Inattentive}} \\ {\large{\bf Utility Maximizer}} };
    \node[align=center] at (5,1.3) { {\large{\bf Equivalent}}\\ {\large {\bf via Theorem \ref{thrm:IRL_DNN}}} };
    \draw [>=triangle 60,thick,->] (ipimg) -- (CNN);
    \draw [>=triangle 60,thick,->] (CNN) -- (opimg);
    \draw[>=triangle 45, thick,double, <->] (bigCNN) --    (RI);
\end{tikzpicture}
\caption{Schematic illustration of rationally inattentive Bayesian utility maximization based interpretable image classification by deep CNNs. Theorem~\ref{thrm:IRL_DNN} establishes equivalence between the image classification behavior of a deep CNN and the decisions of a rationally inattentive maximizer. Hence, the deep CNN's image classification behavior can be parsimoniously represented by a utility function and an information acquisition cost.
}
\label{fig:equivalence}
\end{figure}
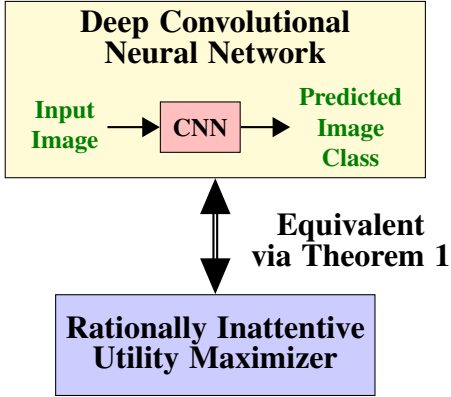

\begin{compactenum}[1.]
\item We show that the image classification decisions of deep CNNs satisfy the necessary and sufficient conditions for rationally inattentive utility maximization by a large margin. The margin by which the decisions satisfy these conditions are displayed in Table~\ref{tab:res_rob_dtest_both}. Hence, we establish that rationally inattentive utility maximization is a robust fit to deep image classification. This result is schematically shown in Fig.\,\ref{fig:equivalence}. 
\item To aid visualization of our interpretable model, we provide a sparsity-enhanced decision test that computes the sparsest utility function and information acquisition cost which rationalizes deep CNN decisions. The sparsest solution yields a parsimonious representation of hundreds of thousands of layer weights of the deep CNNs in terms of a few hundred parameters. The utility function of the sparsest interpretable model also induces a useful preference ordering amongst the set of hypotheses (image labels) considered by the CNN; for example, how much additional priority
is allocated to the classification of a cat as a cat compared
to a cat as a dog. In classical deep learning, this preference
ordering is not explicitly generated. The sparsity results for various deep CNN architectures are displayed in Table~\ref{tab:sparse} and Fig.\,\ref{fig:reconstruct_cost}.
\item Our final result demonstrates the usefulness of our interpretable model. We show that the  interpretable model computed from  CNN decisions can predict the classification accuracy of a  CNN trained with arbitrary parameters with accuracy exceeding 94\%. This by-passes the need to re-train the deep CNN when its accuracy is observed for a finite number of training parameters. The prediction results are displayed in Table~\ref{tab:prederrors}.
\end{compactenum}
The above results are backed by experiments performed on several deep CNN architectures using the benchmark image dataset, namely, CIFAR-10~\cite{KRZ09}. The first two results use deep CNN decisions aggregated over varying training epochs. The third (prediction) result uses deep CNN decisions trained on noisy image datasets parameterized by the noise variance.
 

\subsection*{Related Works} 
Since we study  interpretable deep learning using behavioral and micro- economics, we briefly discuss related works in these areas.

{\em Bayesian revealed preference and Rational inattention.} 
Estimating utility functions given a finite sequence of decisions and budget constraints is the central theme of revealed preference in micro-economics. The seminal work of \cite{AF67,DW73}~(see also \cite{VR82}) give necessary
and sufficient conditions for the existence of a 
utility function that rationalizes a finite time series of consumption bundles of a decision-maker. Rationally inattentive models for Bayesian decision making have been studied extensively in~\cite{WD12,MAT15,DE17}. In the last decade, the area of Bayesian revealed preference~\cite{CM15,CD15} develops necessary and sufficient conditions to test for rationally inattentive Bayesian utility maximization.


{\em Interpretable ML.} 
Providing transparent models for de-obfuscating  `black-box' ML algorithms under the area of interpretable machine learning is  a subject of extensive research~\cite{CHK17,DS17,GD18}. Interpretable machine learning is defined in \cite{MR19} as ``the use of machine-learning models for the extraction of
relevant knowledge about domain relationships contained in
data''. 

Since the literature is enormous, we only discuss a subset of works pertaining to interpretability of deep neural networks for image classification~\cite{CI12,RS15,HE16}. 
One prominent approach, namely, saliency maps,
reconstructs the most preferred or typical image pertaining to each image class the deep neural network has learned~\cite{SM13,NG16}. Related work includes creating hierarchical models for determining the importance of image features that determine its label~\cite{CR19}. In this paper, this feature importance is encoded into the utility function that parameterizes our interpretable model.
Another approach seeks to provide local approximations to the trained model, local w.r.t the input image \cite{LEI16,LB17}. In contrast, our generative interpretable model provides a global black-box approximation for deep image classification. A third approach approximates the decisions of the deep neural networks by a linear function of simplified individual image features~\cite{BC15,RB16,SHR16,LB17}. 
In contrast, our interpretable model fits a stochastic non-linear map that relates the true and predicted image labels. The parameters of the map are obtained by solving a convex feasibility problem parameterized by the deep CNN decisions.
Finally, deep neural networks have also been modeled by Bayesian inference frameworks using probabilistic graphical methods \cite{WN16}.

To the best of our knowledge, an economics based approach for the post-hoc analysis of deep neural networks has not been explored in literature. However, we note that behavioral economics based interpretable models have been applied to domains outside interpretable machine learning, for example, in online finance platforms for efficient advertising \cite{Mil81,HNG07}, training neural networks~\cite{MIR21} and more recently in YouTube to rationalize user commenting behavior~\cite{HKP19}. Finally, due to our recent equivalence result~\cite{PK21}, our behavioral economics approach to interpretable deep image classification can be related to classical revealed preference methods~\cite{AF67,DW73} in microeconomics. 


\section{Bayesian Revealed preference 
with Rational Inattention}\label{sec:foundation}
This section describes the key ideas behind Bayesian revealed preference.
Despite the abstract formulation below, the reader should keep in mind the deep learning context. In Sec.\,\ref{sec:results_basic}, we will use Bayesian revealed preference theory to construct an interpretable  deep learning representation by showing that deep CNNs are equivalent to rationally inattentive Bayesian utility maximizers.


\subsection{Utility Maximization with Rational Inattention (\RImodel)}\label{sec:umri_model}
Bayesian revealed preference aims to determine if the decisions of a Bayesian agent are consistent with expected utility maximization subject to a rational inattention sensing cost. We start by describing the {\bf u}tility {\bf m}aximization model with {\bf r}ational {\bf i}nattention (henceforth called \RImodel) for a \textbf{{\em collection}} of Bayesian decision makers/agents. 

Abstractly, the \RImodel\ model is parameterized by the tuple 
\begin{equation}\label{eqn:RI_tuple}
\bigtuple = (\dpset,\stateset,\obsset,\actionset,\prior,\RIcost,\{\attfunagent{\dpiter},\utilitysymbol_\dpiter,\dpiter\in\dpset\}).
\end{equation}
With respect to the abstract parametrization of the \RImodel\ model for a collection of Bayesian agents, the following elements constitute the tuple $\bigtuple$ defined in  \eqref{eqn:RI_tuple}.\\ 
\underline{{\em Agents:}} $\dpset=\{1,2,\ldots,\numdp\}~(\numdp\geq 2)$ indexes the finite set of Bayesian agents.\\
\underline{{\em State:}} $\stateset$ is the finite set of ground truths with prior probability distribution $\prior$. With respect to our image classification context, $\stateset=\{1,2,\ldots 10\}$ is the set of image classes in the CIFAR-10 dataset and $\prior$ is the empirical probability distribution of the image classes in the test dataset of CIFAR-10. \\
\underline{{\em Observation and attention strategy:}} Agent $\dpiter\in\dpset$ chooses attention strategy $\attfunagent{\dpiter}:\stateset\rightarrow\Delta(\obsset)$,
a stochastic mapping from $\stateset$ to a finite set of observations $\obsset$. Given state $\state$ and attention strategy $\attfunsymb_\dpiter$, the agent samples observation $\obs$ with probability $\attfunagent{\dpiter}(\obs|\state)$. 
The agent then computes the posterior probability distribution $p(\state|\obs)$ via Bayes formula  as
\begin{equation}\label{eqn:compute_posterior}
p(\state|\obs) = \frac{  \prior(\state)\attfunsymb_\dpiter(\obs|\state) }{ \sum_{\state'\in\stateset} \prior(\state')\attfunsymb_\dpiter(\obs|\state')  }.
\end{equation}
The observation and attention strategy are latent variables that abstractly represent the learned feature representations in the deep image classification context. Bayesian revealed preference theory tests their existence via the convex feasibility test in Theorem~\ref{thrm:IRL_DNN} below.\\
\noindent\underline{{\em Action:}} Agent $\dpiter\in\dpset$ 
chooses action $\action$ from a finite set of actions $\actionset$ after computing the posterior probability distribution $p(\state|\obs)$. In the image classification context, $\action$ is the image class predicted by the neural network, hence $\actionset=\stateset$.
\\
\underline{{\em Utility function:}} Agent $\dpiter\in\dpset$ has a utility function $\utilitysymbol_\dpiter(\state,\action)\in\reals^+$, $\state\in\stateset,\action\in\actionset$ and aims to maximize its expected value, with the expectation taken wrt the random state $\state$ and random observation $\obs$. A key feature in our approach is to show that the utility function rationalizes the decisions of the deep CNNs (made precise in Definition~\ref{def:RI}).\\
\underline{{\em Information Acquisition Cost:}} The information acquisition cost $\RIcost(\attfunsymb,\prior)\in\reals^+$ depends on attention strategy $\attfunsymb$ and prior pmf $\prior$. It is the sensing cost the agent incurs in order to
 estimate  the underlying state \eqref{eqn:compute_posterior}. In the context of machine learning, $\RIcost(\cdot)$ abstractly captures the `learning' cost incurred during training of the deep neural networks. In rational inattention theory from behavioral economics, 
a higher information acquisition cost is incurred for more accurate attention strategies (equivalently, more accurate state estimates \eqref{eqn:compute_posterior} given observation $\obs$). 
We refer the reader to the influential work of \cite{Sim03,Sim10}.


Each Bayesian agent $\dpiter\in\dpset$,  aims to maximize its expected utility while minimizing its cost of information acquisition. Hence, the action $\action$ given observation $\obs$, and attention strategy $\attfunsymb_\dpiter$ are  chosen as follows:

\begin{definition}[Rationally Inattentive Utility Maximization] \label{def:RI} Consider a collection of Bayesian agents $\dpset$ parameterized by $\bigtuple$ in \eqref{eqn:RI_tuple} under the \RImodel\ model. Then,\\
\noindent (a) {\bf Expected Utility Maximization:}  
Given posterior probability distribution $p(\state|\obs)$, every agent $\dpiter\in\dpset$ chooses action $\action$ that maximizes its  expected utility. That is, with $\E$ denoting mathematical expectation, the action $\action$ satisfies
\begin{equation}
\action \in\argmax_{\action' \in \actionset}~\E_{\state}\{ \utilitysymbol_\dpiter(\state,\action') | \obs\}= \sum\limits_{\state\in\stateset}p(\state|\obs)\utilitysymbol_\dpiter(\state,\action')
\label{eqn:utilitymaximization}
\end{equation}
\noindent (b) {\bf Attention Strategy Rationality:} For agent $\dpiter$, the attention strategy $\attfunagent{\dpiter}$ optimally trades off between maximizing the expected utility and minimizing the information acquisition cost.
\begin{align}
\attfunsymb_\dpiter \in\argmax_{\attfunsymb'} 
\E_{\obs}\{&\operatorname*{max}_{\action\in\actionset}\E_{\state}\{ \utilitysymbol_\dpiter(\state,\action) | \obs\}\}- \RIcost(\attfunsymb',\prior)
\label{eqn:attentionmaximization}
\end{align}
\end{definition}
Eq.\,\ref{eqn:utilitymaximization},\ref{eqn:attentionmaximization} in Definition~\ref{def:RI} constitute  a nested optimization problem. The lower-level optimization task is to choose the the `best' action for any observation $\obs$ based on the computed posterior belief of the state. The upper-level optimization task is to sample the observations optimally by choosing the `best' attention strategy.

{\em Remark.} The multiple Bayesian agents in $\bigtuple$ have the same state space $\stateset$, observation space $\obsset$, action space $\actionset$, prior $\prior$ and cost of information acquisition $\RIcost$, but only differ in their utility functions. Bayesian revealed preference theory relies on this crucial constraint on the optimization variables in \eqref{eqn:utilitymaximization},~\eqref{eqn:attentionmaximization} for detecting optimal behavior in a finite number of agents.

\subsection{Bayesian Revealed Preference (\dtest) Test for Rationally Inattentive Utility Maximization}
Having described the \RImodel\  model (collection of rationally inattentive utility maximizers), we are now ready to state our key result. Theorem \ref{thrm:IRL_DNN} below says that the decisions of a collection of Bayesian agents is rationalized by a \RImodel\ tuple $\bigtuple$ {\em if and only if} a set of convex inequalities have a feasible solution. 
These inequalities comprise our {\bf B}ayesian {\bf R}evealed {\bf P}reference (henceforth called \dtest) test for rationally inattentive utility maximization.


For notational convenience, the decisions of the Bayesian agents in the \RImodel\ model are  compacted into the dataset $\datasetaccum$ defined as:
\begin{equation} \label{eqn:dataset_accum}
\datasetaccum=\{\prior,p_{\dpiter}(\action|\state),\state\in\stateset,\action\in\actionset,\dpiter\in\dpset\}.
\end{equation} 
In \eqref{eqn:dataset_accum}, $\prior\in\Delta^{|\stateset|-1}$ denotes the prior pmf over the set of states $\stateset$ in $\bigtuple$ \eqref{eqn:RI_tuple}. The variable $p_{\dpiter}(\action|\state)$ is the conditional probability that agent $\dpiter\in\dpset=\{1,2,\ldots,\numdp\}$ takes action $\action$ given state $\state$. $\datasetaccum$ characterizes the input-output behavior of the collection of Bayesian agents and serves as the input for \dtest\ feasibility test described below. 


\begin{theorem}[\dtest\ Test for Rationally Inattentive Utility Maximization \cite{CD15}]\label{thrm:IRL_DNN} Given the dataset $\datasetaccum$ \eqref{eqn:dataset_accum} obtained from a collection of Bayesian agents $\dpset$. Then,\\
{\em 1.} \underline{Existence:}  There exists a \RImodel\ tuple $\bigtuple(\datasetaccum)$ \eqref{eqn:RI_tuple} that rationalizes dataset $\datasetaccum$ if and only if there exists a feasible solution that satisfies the set of convex inequalities 
\begin{equation}\label{eqn:BRP_ineq}
\text{\dtest}(\datasetaccum,\{\utilitysymbol_\dpiter,\runcostinst_\dpiter\}_{\dpiter=1}^{\numdp}) \leq \mathbf{0},~\utilitysymbol_\dpiter\in\reals_+^{|\stateset|\times|\actionset|},~\runcostinst_\dpiter>0.
\end{equation}
In \eqref{eqn:BRP_ineq},~\dtest$(\cdot)$ corresponds to a set of convex (in the variables $\{\utilitysymbol_\dpiter,\runcostinst_\dpiter\}_{\dpiter=1}^{\numdp}$) inequalities, stated in Algorithm~\ref{alg:dtest}. \\
{\em 2.} \underline{Reconstruction:} Given a feasible solution $\{\utilitysymbol_\dpiter,\runcostinst_\dpiter\}_{\dpiter=1}^\numdp$ to $\dtest(\datasetaccum,\cdot)$, $\utilitysymbol_\dpiter$ is the $\dpiter\uth$ Bayesian agent's utility function in the feasible model tuple $\bigtuple(\datasetaccum)$. The set of observations $\obsset=\actionset$, the set of actions in $\datasetaccum$. The feasible cost of information acquisition $\RIcost$ in $\bigtuple(\datasetaccum)$ is defined in terms of $\runcostinst_\dpiter$ as:  
\begin{align}
    \RIcost(\attfunsymb) & = \max_{\dpiter\in\dpset} \runcostinst_\dpiter + \sum_{\action}\max_{\actiontwo\in\actionset}\sum_{\state} p(\state,\action)\utilitysymbol_\dpiter(\state,\actiontwo)\nonumber \\
    & \quad \quad \quad ~- \sum_{\state,\action}p_{\dpiter}(\state,\action)\utilityagent{\dpiter},~\attfunsymb=\{p(\action|\state)\}.\label{eqn:BRP_reconstruct}
\end{align}
\end{theorem}
\noindent The proof of Theorem~\ref{thrm:IRL_DNN} is in Appendix~\ref{proof:Thrm_BRP}.
Before launching into a detailed discussion,
we stress the ``iff" in Theorem \ref{thrm:IRL_DNN}. Put simply:
if the inequalities in \eqref{eqn:BRP_ineq} are not feasible, then the Bayesian agents that generate the dataset $\datasetaccum$ are not rationally inattentive utility maximizers.  
If \eqref{eqn:BRP_ineq} has a feasible solution, then there exists a reconstructable family of viable utility functions and information acquisition costs that rationalize $\datasetaccum$\footnote{In terms of interpretable deep learning, of all parameters
in the \RImodel\ tuple, we are only interested in the utility functions of the agents and the cost of information
acquisition, since the remaining parameters are immediately deduced from the decision dataset $\datasetaccum$.}. A key feature of Theorem~\ref{thrm:IRL_DNN} is that the estimated utilities (and information costs) are set-valued; every utility and cost function in the feasible set explains $\datasetaccum$ equally well. The estimated \RImodel\ model parameters are set-valued due to the finite number of Bayesian agents whose decisions constitute the dataset $\datasetaccum$. The estimated parameter set converges to a point if and only if the inequality \eqref{eqn:BRP_ineq} holds as $|\dpset|\rightarrow\infty$.

{\em Computational Aspects of \dtest\ Test.} Suppose the dataset $\datasetaccum$ is obtained from $\numdp$ Bayesian agents. Then, \dtest$(\datasetaccum)$ comprises
a feasibility test with $\numdp~(|\stateset||\actionset|+1)$ free variables and
$\numdp^2 + \numdp~(|\actionset|^2-|\actionset|-1)$ convex inequalities. Thus, the number of free variables and inequalities in the \dtest\ feasibility test scale linearly and quadratically, respectively, with the number of observed Bayesian agents.\\
 

\begin{algorithm}[ht]
\begin{algorithmic}
\REQUIRE Dataset $\datasetaccum=\{\prior,p_{\dpiter}(\action|\state),\state,\action\in\stateset,\dpiter\in\dpset\}$ from a collection of Bayesian agents $\dpset$.

\hspace{-0.35cm}\textbf{Find:} Positive reals $\runcostinst_{\dpiter},~\utilityagent{\dpiter}\in(0,1]$ for all $\state\in\stateset,$ $\action\in\actionset,~\dpiter\in\dpset$ that satisfy the following inequalities:
\begin{align}
    \hspace{-1.5cm}\underline{\textbf{NIAS}}:&~\sum_{\state}p_{\dpiter}(\state|\action)~(\utilitysymbolagent{\dpiter}(\state,\actiontwo) -\utilityagent{\dpiter})\leq 0,\label{eqn:NIAS}\\
    ~&~ \forall\action,\actiontwo\in\actionset,~\dpiter\in\dpset, \nonumber\\ 
    \hspace{-1.5cm}\underline{\textbf{NIAC}}:&~\sum_{\action}\left( \max_{\actiontwo}\sum_{\state}p_{\dpitertwo}(\state,\action)\utilitysymbolagent{\dpiter}(\state,\actiontwo) \right)  - \runcostinst_{\dpitertwo}\label{eqn:NIAC}\\
    -&~\sum_{\state,\action}p_{\dpiter}(\state,\action)\utilityagent{\dpiter} +\runcostinst_{\dpiter}\leq 0,~\forall\dpitertwo,\dpiter\in\dpset,\nonumber
\end{align}
\hspace{-0.14cm}where $p_{\dpiter}(\state,\action)=\prior(\state)p_{\dpiter}(\action|\state),~p_{\dpiter}(\state|\action)=\frac{p_{\dpiter}(\state,\action)}{\sum_{\state'}p_{\dpiter}(\state',\action)}$.\\\vspace{0.1cm}

\hspace{-0.35cm}\textbf{Return:} Set of feasible utility functions $\utilitysymbolagent{\dpiter}$ and information acquisition costs $\runcostinst_{\dpiter}$ incurred by agents $\dpiter\in\dpset$.
\end{algorithmic}
\caption{\dtest\ Convex Feasibility Test of Theorem~\ref{thrm:IRL_DNN}}
\label{alg:dtest}
\end{algorithm}
\subsection{Relating \RImodel\ and \dtest\ test to Interpretable Deep Image Classification}\label{sec:BRP_insight}

We now discuss how the above \dtest\ test relates to interpretable image classification using deep CNNs.
The \dtest\ convex feasibility test in Theorem~\ref{thrm:IRL_DNN} comprises two sets of inequalities, namely, the {\em NIAS} (No-Improving-Action-Switches) \eqref{eqn:NIAS} and {\em NIAC} (No-Improving-Action-Cycles) \eqref{eqn:NIAC} inequalities (Algorithm~\ref{alg:dtest}). NIAS ensures that the agent takes the best action given a posterior pmf. NIAC ensures that every agent chooses the best attention strategy. \dtest\ test checks if there exist $\numdp$ utility functions and $\numdp$ positive reals that, together with $\datasetaccum$, satisfy the NIAS and NIAC inequalities.

\subsubsection*{Toy Example with 2 CNNs} The following discussion   gives additional insight into our approach.
Consider the simplest case involving two trained deep CNNs $N_1$ and $N_2$; so  $\dpset=\{1,2\}$ in the above notation. Assume $N_1$ and $N_2$ have the same network architecture. Suppose an analyst observes that $N_1$ makes accurate decisions on a rich input image dataset while $N_2$ makes less accurate decisions on the same dataset. 

Our \RImodel\ model first abstracts the accuracy of the feature representations of the input image data learned by $N_1$ and $N_2$ via attention strategies $\attfunsymb_1$ and $\attfunsymb_2$ in~\eqref{eqn:attentionmaximization}. Second, the information acquisition cost function $\RIcost(\cdot)$ abstracts the computational resources expended for learning the representations. The rationale is that learning an accurate latent feature representation is costly, and this  is abstracted by the  information acquisition cost.

Let the training cost incurred by $N_1$ and $N_2$ be $\RIcost(\attfunsymb_1)$ and $\RIcost(\attfunsymb_2)$ respectively. 
If the decisions of $N_1$ and $N_2$ can be explained by the \RImodel\ model (and Theorem \ref{thrm:IRL_DNN} below will give necessary and sufficient conditions for this), then there exist utility functions $\utilitysymbol_1$ and $\utilitysymbol_2$ for $N_1$ and $N_2$,  that satisfy: 
\begin{equation}
    \mathbb{E}_{\attfunsymb_i}\{\utilitysymbolagent{i}\} - \RIcost(\attfunsymb_i) \geq \mathbb{E}_{\attfunsymb_j}\{\utilitysymbolagent{i}\} - \RIcost(\attfunsymb_j),~i,j\in\{1,2\}~\label{eqn:toy_NIAC}
\end{equation}
The above inequality says that  CNNs $N_1$ and $N_2$ would be worse off (in an expected utility sense)  if they make decisions based on swapping each other's learned representations. That is, both $N_1$ and $N_2$ learn the `best' feature representation of the input images given their training parameters. 

\subsubsection*{Discussion}
(i) {\em Parsimonious Interpretable Representation of deep CNNs.} In the deep image classification context, due to the  \RImodel\ model's parsimonious parametrization in  \eqref{eqn:RI_tuple}, the decisions of $\numdp$ CNNs can be rationalized by just $\numdp$ utility functions and an information acquisition cost function, thus bypassing the need of several million parameters to describe the deep CNNs. \\
(ii) {\em Identifiability.}
The \dtest\ feasibility test requires the dataset $\datasetaccum$ to be generated from $\numdp>2$ Bayesian agents. If $\numdp=1$, then \eqref{eqn:BRP_ineq} holds trivially since any information acquisition cost satisfies the convex inequalities of \dtest. Another intuitive way of motivating a collection of agents for the \dtest\ is as follows. Reconstructing a feasible \RImodel\ model tuple $\bigtuple$ that rationalizes the decisions of the deep CNNs is analogous to fitting a line to a finite number of points. One can fit infinitely many lines through a single point. The task becomes non-trivial if the number of points exceeds $2$. In the Bayesian revealed preference context, the points correspond to the decisions from each Bayesian agent. The slope and intercept of the fitted line, in our case, corresponds to the utility function and cost of information acquisition that rationalize the agent decisions.
\\
(iii) {\em Relative Optimality implies Global Optimality.} In the setting involving $\numdp>2$ deep CNNs (agents), the NIAS and NIAC inequalities of \dtest\ test check for relative optimality - 
{\em given utility function $\utilitysymbol_\dpiter$, does deep CNN $\dpiter$ performs at least as well as any other observed deep CNN in $\dpset\backslash\{\dpiter\}$?} Clearly, testing for relative optimality is weaker than testing for global optimality~\eqref{eqn:attentionmaximization} which ideally requires access to decisions from an infinite number of deep CNNs. Setting the cost of information acquisition as a free variable bridges this gap.
The proof of Theorem~\ref{thrm:IRL_DNN} shows that if the deep CNN decisions satisfy relative optimality, then there exists a cost of information acquisition such that the decisions are globally optimal. That is, Theorem~\ref{thrm:IRL_DNN} ensures relative optimality is sufficient for global optimality.\\
(iii) {\em Generalization of \cite{CD15}.} Theorem~\ref{thrm:IRL_DNN} generalizes \cite[Theorem 1]{CD15} in two ways. (1) In \cite{CD15}, the utilities $\utilitysymbolagent{\dpiter}$ in \RImodel\ model tuple $\bigtuple$ are assumed known, and only the information acquisition costs $\runcostinst_{\dpiter}$ are estimated, whereas Theorem~\ref{thrm:IRL_DNN} estimates both parameters. (2) The expression for the reconstructed model tuple $\bigtuple(\datasetaccum)$ is novel; the discussion in \cite{CD15} is only confined to the existence of such a tuple.\\
(vi) {\em Single Utility \RImodel\ (\RImodels\,).} In Appendix~\ref{sec:umri_model_compact}, we propose a sparse version of \RImodel\,, namely, the \RImodels\ model in \eqref{eqn:RI_tuple_compact}. The key distinction of this model is that all agents have the same utility function $\utilitysymbol$ and thus can be represented with substantially fewer parameters. In complete analogy to Theorem~\ref{thrm:IRL_DNN}, we outline a decision test in Theorem~\ref{thrm:IRL_DNN_compact} that states necessary and sufficient conditions for agent decisions to be consistent with the \RImodels\ model of rationally inattentive utility maximization. We discuss this sparse parametrization in  the appendix so as not to interrupt the flow  of  the main text.\\
(vii) {\em Degenerate solution to \dtest\ and \dtestcompact\ tests.} The degenerate utility function of all zeros and cost of information acquisition $\RIcost=0$ trivially satisfy the \dtest\ and \dtestcompact\ tests and lie at the boundary of the feasible set of parameters.\\

\subsubsection*{Summary}
This section formulated an economics-based decision-making model. Since this model may not be familiar to a machine learning reader, we  summarize the main ideas.
We introduced the rationally inattentive utility maximization model, namely, the \RImodel\ model for a collection of Bayesian agents (decision makers). Our main result  Theorem~\ref{thrm:IRL_DNN}  outlines a decision test \dtest\ for rationally inattentive utility maximization given decisions from a collection of agents. This \dtest\ test comprises a set of convex inequalities that have a feasible solution {\em if and only if} the collection of agents are rationally inattentive utility maximizers. Theorem~\ref{thrm:IRL_DNN} also provides an explicit reconstruction of the feasible \RImodel\ model parameters that rationalize input agent decisions. The set of feasible utility functions and information acquisition costs thus parsimoniously explain the decisions generated by the Bayesian agents. In Appendix~\ref{sec:umri_model_compact}, we propose a single utility version of the \RImodel\ model with fewer parameters. Due to fewer parameters, the decision test for this sparse model, given in  Theorem~\ref{thrm:IRL_DNN_compact}, is computationally less expensive yet more restrictive than the \dtest\ test for rationality in Theorem~\ref{thrm:IRL_DNN}. 


The rest of the paper focuses on computing interpretable \RImodel\ models that rationalize deep CNN decisions. 
We will investigate through extensive experiments how well the \RImodel\ fits the deep CNN decisions via robustness tests. We will also investigate how well the computed interpretable models, namely, \RImodel\ and \RImodels\,, predict the deep CNNs' decisions when the training parameters are varied.
\section{Bayesian Revealed Preference explains CIFAR-10 Image Classification by Deep CNNs} \label{sec:results_basic}
The  experimental results in this section are divided into two parts: First, we show that the deep CNNs decisions pass the \dtest\ and \dtestcompact\ tests formulated in Theorems~\ref{thrm:IRL_DNN} and \ref{thrm:IRL_DNN_compact} by a large margin. This implies  
 that the rationally inattentive utility maximization model is a robust fit to the deep CNN decisions.

Our second result demonstrates an   application of the reconstructed interpretable model. Training datasets are often noisy. We show that in such a noisy setting, the reconstructed interpretable model from Theorem~\ref{thrm:IRL_DNN} can accurately predict (with accuracy exceeding 94\%) the image classification performance of the deep CNNs. This bypasses the need to train the deep CNN for various noise variances that corrupt the training dataset. 

\subsection*{Experimental Setup: Deep CNN Architectures, Training Parameters and Construction of Dataset}
\label{sec:nn_model_description}

{\em Image Dataset.} In our experiments, we trained and validated the deep CNNs using the CIFAR-10 benchmark image dataset~\cite{KRZ09}.
This  public dataset  consists of $60000$ $32$x$32$ colour images in $10$ distinct classes (for example, airplane, automobile, ship, cat, dog etc.), with $6000$ images per class. There are $50000$ training images and $10000$ test images. We will use the terms image classes and image labels interchangeably.\footnote{Our experiments are confined to the CIFAR-10 dataset for clarity of exposition. Our approach to interpretable deep learning can be easily extended to richer benchmark image datasets like ImageNet and CIFAR-100 (that comprise over a 100 image labels).}

{\em Network Architecture and Training Parameters.} In this paper, we use $5$ well-known deep CNN architectures for our experiments.
1. LeNet~\cite{LC89},
2. AlexNet~\cite{KRZ12}
3. VGG16~\cite{SMN14}
4. ResNet-50~\cite{HE16}
5. Network-in-Network (NiN)~\cite{LIN13}
 The deep CNNs are trained and validated on the CIFAR-10 image dataset, using $3$ learning rate schedules, namely, L.R.\ 1, L.R.\ 2 and L.R.\ 3. All $3$ schedules use the RMSprop optimizer \cite{HN12} with the decay parameter and maximum training epochs (full passes of the training dataset) set to $10^{-6}$ and $200$, respectively, and initial step size set to $0.01$. The step size is halved every $20,30,40$ epochs, respectively, for L.R.\ 1, 2 and 3.

{\em Relation to Bayesian revealed preference.} We now relate the deep CNN setup to the Bayesian revealed preference framework in Sec.\,\ref{sec:foundation}.
For each CNN architecture, we use the decisions of $\numdp=20$ CNNs, i.e.\,, $20$ Bayesian agents in the terminology of Sec.\,\ref{sec:foundation}, for our \dtest\ and \dtestcompact\ decision tests. The CNN decisions from $\numdp$ CNNs on the test image dataset of CIFAR-10 are aggregated into dataset $\datasetaccum$~\eqref{eqn:dataset_accum}. The results of the decision tests are discussed below. In the deep image classification context, the parameter $p_\dpiter(\action|\state)$ in~\eqref{eqn:dataset_accum} is the probability that the $\dpiter\uth$ deep CNN classifies an image from category $\state$ into category $\action$ in the CIFAR-10 test image dataset. The prior $\prior$ in $\datasetaccum$~\eqref{eqn:dataset_accum} is the empirical pmf over the set of image categories in the CIFAR-10 test dataset. 
Constructing $\datasetaccum$ from raw CNN decisions is discussed in  Appendix~\ref{sec:construct_dataset}.

\subsection{\dtest\ and \dtestcompact\ Tests for deep CNN datasets. Results and Insights} \label{sec:results_epochs}
\begin{table}[h]
\centering
    \begin{tabular}{|m{0.35\linewidth}|m{0.1\linewidth}|m{0.1\linewidth}|m{0.1\linewidth}|}
    \hline    
    \centering
     Network Architecture & Learning Rate & $\robmet$ $(\times10^{-4})$ & $\robmetcompact$ $(\times10^{-4})$ \\ \hline
    \centering\multirow{3}{*}{LeNet}& L.\,R.\,~1& $30.34$  & $4.72$ \\ 
     & L.\,R.\,~2 & $35.14$  & $4.65 $\\
     & L.\,R.\,~3 & $37.97$  & $5.11$\\ \hline
     \centering\multirow{3}{*}{AlexNet}& L.\,R.\,~1& $32.10$  & $3.21$\\
     & L.\,R.\,~2 & $34.98$  &  $3.91$\\
     & L.\,R.\,~3 & $40.60$ &  $4.62$ \\ \hline
     \centering\multirow{3}{*}{VGG16}& L.\,R.\,~1&  $96.36$  & $4.09$\\
     & L.\,R.\,~2 & $107.4$  & $4.02$\\
     & L.\,R.\,~3 & $119.8$  & $4.44$\\ \hline
     \centering\multirow{3}{*}{ResNet-50}& L.\,R.\,~1& $126.2$  & $2.82$\\
     & L.\,R.\,~2 & $129.2$  & $3.45$\\
     & L.\,R.\,~3 & $132.3$  & $3.83$\\ \hline
     \centering\multirow{3}{*}{Network-In-Network (NiN)}& L.\,R.\,~1& $108.3$  & $3.59$\\
     & L.\,R.\,~2 & $132.1$  & $3.36$\\
     & L.\,R.\,~3 & $149.1$  & $5.57$\\ \hline
    \end{tabular}
    \caption{How  does increasing the number of degrees of freedom of the interpretable model improve robustness of fit to the CNN decisions?
    We see that $\robmet$~\eqref{eqn:margin_dtest} is substantially higher (by an order of magnitude) than $\robmetcompact$~\eqref{eqn:margin_dtest_compact} for all CNN architectures.
    We conclude that the \RImodel\ model fits CNN decisions substantially better than the \RImodels\ model, but with larger computing cost for evaluating the parameters of the interpretable model. Thus, if there are no computational constraints, we recommend using the \RImodel\ model for interpreting CNN decisions.
    }
    \label{tab:res_rob_dtest_both}
\end{table}
\subsubsection*{A. Robustness Results on Deep CNN datasets}
Our first key result is that image classifications of all $5$ deep CNN architectures listed in Sec.\,\ref{sec:nn_model_description} pass the \dtest\ and \dtestcompact\ tests by a large margin. The results are tabulated in Table~\ref{tab:res_rob_dtest_both}. The robustness values $\robmet$ and $\robmetcompact$ in Table~\ref{tab:res_rob_dtest_both} are defined in Definition~\ref{def:margin} below which formalizes the notion of margin for the decision tests. 

\begin{table*}
    \centering
    \begin{tabular}{|m{0.16\linewidth} |m{0.07\linewidth}  |m{0.045\linewidth}|m{0.04\linewidth}|m{0.04\linewidth}|m{0.04\linewidth}|m{0.04\linewidth}|m{0.04\linewidth}|m{0.04\linewidth}|m{0.04\linewidth}|m{0.04\linewidth}|m{0.04\linewidth}|}
    \hline
    \centering
     Network Architecture & Learning Rate (L.R.) &  airplane & auto & bird & cat & deer & dog	& frog	& horse & ship & truck\\ \hline 
     \centering\multirow{3}{*}{LeNet} & L.R.\ 1 & $17.61$ & $3.55$ & $20.06$ & $1.88$ & $17.19$ & $21.42$ & $42.00$ & $27.79$ & $1.91$ & $9.55$ \\ 
     & L.R.\ 2 & $4.13$ & $5.20$ & $7.82$ & $1.90$ & $13.18$ & $18.66$ & $23.84$ & $8.16$ &  $2.48$ & $2.47$ \\ 
      & L.R.\ 3 & $10.79$  & $8.27$ & $18.62$ & $22.67$    &  $19.91$    &  $25.01$    &  $47.71$    &  $73.52$    &  $2.65$ &  $1.01$\\ \hline
      \centering\multirow{3}{*}{AlexNet} & L.R.\ 1 & $210.78$ & $41.84$ & $49.77$ & $59.71$ & $51.24$ & $68.31$ & $83.94$ & $211.61$ & $60.43$ & $125.73$ \\
     & L.R.\ 2 & $85.51$ & $47.89$ & $17.38$ & $1.00$ & $25.34$ &  $202.78$ & $21.30$ & $35.01$ & $533.62$ & $248.57$ \\
      & L.R.\ 3 & $18.00$ & $49.55$ & $58.25$ & $28.31$ & $135.54$ & $29.24$ & $224.91$ & $214.51$ & $8.29$ & $264.20$\\ \hline
      \centering\multirow{3}{*}{VGG16} & L.R.\ 1 & $164.48$ & $154.77$ & $15.42$ & $33.67$ & $6.28$ & $123.89$ & $62.83$ & $26.21$ & $1.43$ & $170.69$ \\ 
     & L.R.\ 2 & $88.73$ & $154.10$ & $45.63$ & $297.61$ & $131.08$ & $136.52$ & $57.34$ & $229.80$ & $145.99$ & $11.90$ \\ 
     & L.R.\ 3 & $24.33$ & $10.78$ & $93.90$ & $11.11$ & $91.96$ & $56.64$ & $77.30$ & $110.60$ & $20.28$ & $17.09$ \\ \hline
     \centering \multirow{3}{*}{ResNet-50} & L.R.\ 1 & $50.83$ & $17.55$ & $16.09$ & $4.66$ & $17.92$ & $3.67$ & $4.92$ & $3.95$ & $15.46$ & $4.88$\\
     & L.R.\ 2 & $7.51$ & $8.40$ & $72.70$ & $30.72$ & $32.43$ & $83.65$ & $221.27$ & $74.59$ & $99.04$ & $20.51$\\
     & L.R.\ 3 & $14.61$ & $367.59$ & $31.61$ & $9.20$ & $16.35$ & $11.58$ & $41.44$ & $243.95$ & $222.67$ & $483.91$\\ \hline
     \centering \multirow{3}{*}{Network-in-Network} & L.R.\ 1 & $5.02$ & $30.95$ & $9.91$ & $71.38$ & $63.69$ & $45.88$ & $31.39$ & $67.86$ & $17.03$ & $21.41$ \\ 
     & L.R.\ 2 & $40.17$ & $60.32$ & $4.40$ & $55.67$ & $95.02$ & $88.72$ & $91.15$ & $15.98$ & $176.75$ & $10.27$ \\
     & L.R.\ 3 & $10.47$ & $75.32$ & $55.97$ & $24.17$ & $17.41$ & $8.94$ & $23.02$ & $71.27$ & $29.94$ & $80.91$ \\ \hline
    \end{tabular}
    \caption{The utility function of the sparsest interpretable model is a diagonal matrix. The diagonal elements yield a natural preference ordering amongst the set of image classes (classification hypotheses). 
    For example, consider the VGG16 architecture trained using learning rate $1$ (third row, first sub-row of table). The maximum utility is for trucks ($170.69$, last column) and the minimum is for ships ($1.43$, second last column). This shows the sparsest interpretable model induces the following preference ordering for the VGG16 architecture: classifying trucks correctly is prioritized $100$ times more than classifying ships. Such a  preference ordering is not explicitly generated by a CNN. 
    }
    \label{tab:sparse}
\end{table*}


\begin{definition}[Robustness (Goodness-of-fit) of \dtest\ and \dtestcompact\ Tests.]\label{def:margin}
 Given dataset $\datasetaccum$~\eqref{eqn:dataset_accum} aggregated from a collection of Bayesian agents, $\robmet(\datasetaccum)$ and $\robmetcompact(\datasetaccum)$ measure the largest perturbation so that $\datasetaccum$ passes the \dtest\ and \dtestcompact\ decision tests:
\begin{align}
    &\robmet(\datasetaccum)  = \max_{\eps>0}\frac{\eps~\numdp}{\sum_{\dpiter=1}^{\numdp}\|\utilitysymbolagent{\dpiter}\|_2^2},~\text{\dtest}(\datasetaccum,\{\utilitysymbol_\dpiter,\runcostinst_\dpiter\}_{\dpiter=1}^\numdp) \leq -\eps.\label{eqn:margin_dtest}\\
    &\robmetcompact(\datasetaccum) = \max_{\eps>0}\frac{\eps}{\|\utilitysymbol\|_2^2}, \text{\dtestcompact}(\datasetaccum,\utilitysymbol,\{\runcostinst_\dpiter,\lambda_\dpiter\}_{\dpiter=1}^\numdp) \leq  -\eps.\label{eqn:margin_dtest_compact}
\end{align}
\end{definition}
\noindent In Definition~\ref{def:margin}, robustness values $\robmet$ and $\robmetcompact$ measure, respectively, the smallest perturbation needed for $\datasetaccum$ to fail the \dtest\ and \dtestcompact\ decisions tests. Both $\robmet$ and $\robmetcompact$ are normalized wrt the row-wise $\mathcal{L}_2$ norm of the feasible utility functions. 
Higher robustness values imply a better fit of the \RImodel\,, \RImodels\ models to the decision dataset \footnote{The robustness value for the non-informative dataset of uniformly distributed pmfs is $0$. Hence, the robustness value measures the informativeness of the attention strategies in $\datasetaccum$ relative to the uniform probability distribution.}. \vspace{0.2cm}


\subsubsection*{Discussion and Insights. Robustness Results of Table~\ref{tab:res_rob_dtest_both}}
(i) {\em Deep CNN dataset:} The deep CNN datasets used for the robustness tests~\eqref{eqn:margin_dtest},~\eqref{eqn:margin_dtest_compact} comprise decisions of $\numdp=20$ deep CNNs for every network architecture, where CNN $\dpiter$ was trained for $10~\dpiter$ training epochs,~$\dpiter=1,2,\ldots,\numdp$.\\
(ii) {\em Comparison between $\robmet$ and $\robmetcompact$ values for deep CNN datasets:} The average value of $\robmetcompact$~\eqref{eqn:margin_dtest_compact}  over all $3$ learning rate schedules and $5$ network architectures was found to be $4.09\times10^{-4}$. In contrast, the average value of $\robmet$~\eqref{eqn:margin_dtest} was found to be $87.45\times10^{-4}$, almost $20$ times the average value of $\robmetcompact$. This result shows that the \RImodel\ model fits deep CNN decisions substantially better than the \RImodels\ model. This result is expected since \RImodels\ is parameterized using much fewer variables compared to the \RImodel\ and hence, \dtestcompact\ test is more restrictive than \dtest\,.\\
(iii) {\em Sensitivity of $\robmet,\robmetcompact$ to Network Architecture:} The average value of $\robmet$ is $122.29\times10^{-4}$ for the LeNet and AlexNet architectures, which is approximately $3.5$ times the the average value of $\robmet$ for the VGG16, ResNet-50 and NiN architectures which is $35.18\times10^{-4}$. The variation of $\robmetcompact$ with network architecture is negligible compared to $\robmetcompact$. This shows the robustness test for \RImodel\ model is more sensitive to network architecture compared to  that for the \RImodels\ model. \\
(iv) {\em Computational aspects of $\robmet$ and $\robmetcompact$.} The computation time for $\robmet$ is almost $30$ times that for $\robmetcompact$. This is expected since the \RImodel\ model is parameterized by $\numdp$ utility functions compared to a single utility function in \RImodels\,.


\subsubsection*{B. Sparsity-enhanced Interpretable Model} 
Our next task is to determine the sparsest possible interpretable model that satisfies the decision tests  \dtest\ and \dtestcompact.  The motivation is three fold:
\begin{compactenum}
\item The sparsest interpretable model explains the deep CNN decisions  using the fewest  number of parameters. 
\item The sparsest interpretable model induces a useful preference ordering amongst the set of hypotheses (image labels) considered by the CNN; for example, how much additional priority
is allocated to the classification of a cat as a cat compared
to a cat as a dog. In classical deep learning, this preference
ordering is not explicitly generated. 
\item Third, the sparsest solution is  a point valued estimate. Recall the \dtest\ and \dtestcompact\ decision tests  yield a set-valued estimate of feasible utility functions and cost of information acquisition that explain the deep CNN datasets. While every element in the set explains the dataset equally well, it is useful to have a single representative point.
\end{compactenum}

Theorem~\ref{thrm:sparse} below computes the sparsest utility function out of all feasible utility functions.

\begin{theorem}[Sparsity Enhanced \dtest\ and \dtestcompact\ Tests for Deep CNN datasets] \label{thrm:sparse} 
Given dataset $\datasetaccum$~\eqref{eqn:dataset_accum} from a collection of $\numdp$ Bayesian agents. The sparsest solutions to the \dtest\ and \dtestcompact\ tests minimize the sum of row-wise $\mathcal{L}_1$ norm of the feasible utility functions of the $\numdp$ agents that generate $\datasetaccum$. 
\begin{align}
& (\utilitysymbol_{1:\numdp})^\ast  =  \argmin_{\utilitysymbolagent{1:\numdp}} \sum_{\dpiter=1}^{\numdp} \norm{\utilitysymbolagent{\dpiter}}_1,\text{\dtest}(\datasetaccum,\cdot)\leq\mathbf{0},~\sum_{\dpiter=1}^\numdp \|\utilitysymbol_\dpiter\|_2^2  = \numdp.\nonumber\\
& \utilitysymbol^\ast  = \argmin_{\utilitysymbol}  \norm{\utilitysymbol}_1,\text{\dtestcompact}(\datasetaccum,\cdot)\leq\mathbf{0},~\|\utilitysymbol\|_2^2 = 1.\label{eqn:sparse_dtest_both}
\end{align}
where $\norm{\cdot}_1$ denotes the row-wise $\mathcal{L}_1$ norm.
\end{theorem}

\subsubsection*{Results and Discussion. Sparsity Test for deep CNN datasets}
\begin{figure}[h]
    \centering
    \includegraphics[width=0.9\linewidth]{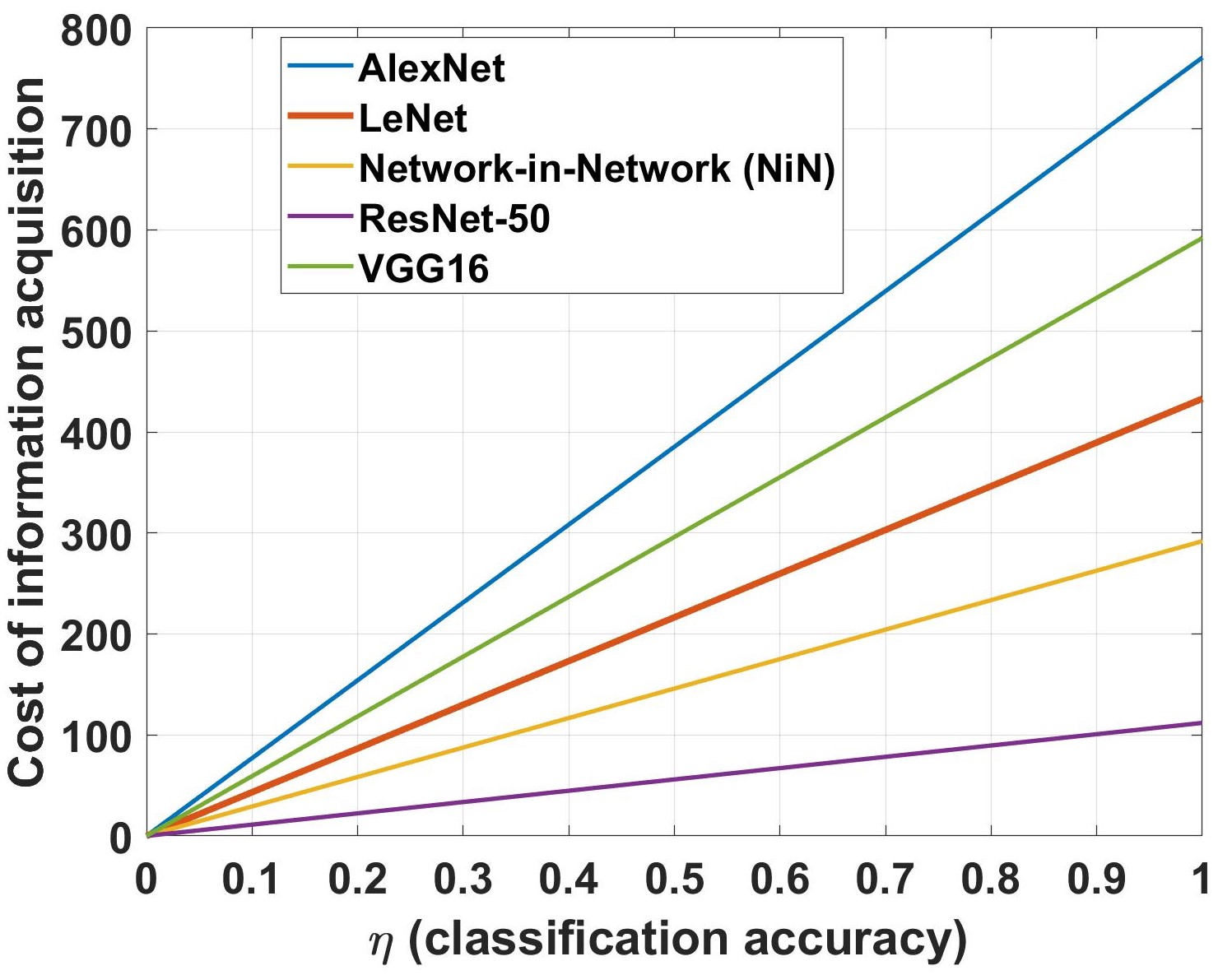}
    \caption{
    The  figure illustrates an important property of our approach to interpretable deep learning: in addition to the utility function (Table~\ref{tab:sparse}), we also need a rational inattention term (cost of learning latent image features) to explain CNN decisions. Put differently, we cannot explain the variation in CNN classification accuracy versus variation in training parameters without an information acquisition cost.    
    The figure displays the information acquisition cost $\RIcost$~\eqref{eqn:BRP_reconstruct} evaluated for the sparsest interpretable model. We also observe that
    learning accurate image features is most expensive for AlexNet, and least expensive for ResNet architectures.
    }
    \label{fig:reconstruct_cost}
\end{figure}

The sparsest utility function from the \dtestcompact\ test are tabulated in Table~\ref{tab:sparse} for all $5$ deep CNN architectures. The corresponding information acquisition cost for all $5$ architectures averaged over learning rates $1,2,3$ are shown in Fig.\,\ref{fig:reconstruct_cost}. Together, the sparsest utility and information cost constitute the sparsest \RImodels\, interpretable model\footnote{For brevity, we have only included the sparsity results for the \RImodels\ model. The sparsest utility functions of the \RImodel\ model that explains deep CNN decisions are included in our public GitHub repository that contains all test results and codes.} for the deep CNN decisions.\\
(i) The sparsest \RImodels\ model is comprised of $\numdp(|\stateset||\actionset|+1)$ variables.  
(ii) {\em Preference ordering induced from sparsest utility.} The sparsest utility function for the \RImodels\ model induces a useful preference ordering among the predicted image classes. That is, they measure how the deep CNN's priority for accurate classification varies across image classes. 
For instance, consider the VGG16 architecture trained using learning rate schedule $1$. Of all image categories, the maximum utility is observed for trucks ($170.69$) and the minimum for ships ($1.43$). This shows the VGG16 architecture prioritizes classifying trucks correctly about $100$ times more than classifying ships.\\
(ii) {\em Penalty for learning image features accurately.} The computed information acquisition costs in Fig.\,\ref{fig:reconstruct_cost} can be understood as the training cost the CNN incurs to learn latent image features accurately. The interpretable model cannot explain the variation in CNN classification accuracy versus variation in training parameters without an information acquisition cost. 
From Fig.\,\ref{fig:reconstruct_cost}, we can conclude that learning accurate image features is the most and least costly, respectively, for the AlexNet and ResNet architectures, respectively. 

\begin{table*}
\centering
    \begin{tabular}{|m{0.2\linewidth} |m{0.045\linewidth}|m{0.04\linewidth}|m{0.04\linewidth}|m{0.04\linewidth}|m{0.04\linewidth}|m{0.04\linewidth}|m{0.04\linewidth}|m{0.04\linewidth}|m{0.04\linewidth}|m{0.04\linewidth}|}\hline \centering
    Network Architecture &  airplane & auto & bird & cat & deer & dog & frog & horse & ship & truck\\ \hline 
    \centering LeNet & $0.042$ & $0.042$ & $0.041$ & $0.027$ & $0.046$ & $0.025$ & $0.049$ & 	$0.034$ & 	$0.040$ & 	$0.042$  \\ \hline 
    \centering AlexNet & $0.025$ & $0.031$ & $0.034$ & 	$0.021$ & 	$0.046$ & $0.032$ & $0.049$ & 	$0.039$ & 	$0.045$ & $0.036$  \\ \hline 
    \centering VGG16 &
    $0.033$ & $0.035$ & $0.043$ & 	$0.041$ & $0.048$ &	$0.048$ & 	$0.035$ & 	$0.046$ & $0.037$ & $0.048$ \\ \hline 
    \centering ResNet-50 & $0.030$ 	& $0.031$ & 	$0.027$	& $0.031$ & $0.020$	& $0.027$	& $0.040$	& $0.015$ & $0.023$ & $0.024$ \\ \hline 
    \centering Network-in-Network &
    $0.051$ &	$0.029$ & 	$0.025$ & $0.028$ & 	$0.056$ & 	$0.059$ & 	$0.030$ & 	$0.058$ & 	$0.045$ & 	$0.036$ \\ \hline 
    \end{tabular}
    \caption{How well does our interpretable model predict CNN classification accuracy? The table displays the prediction error $\delta_\eta(\state)$ defined in~\eqref{eqn:pred_error_def}. Recall $\delta_\eta(\state)$ is the error between the true CNN performance and the predicted performance using the interpretable model with Algorithm~\ref{alg:pred_interpolate}.
    The maximum error across all image classes and architectures was found to be $5.9\%$. Hence, our interpretable model predicts CNN classification performance with accuracy exceeding $94\%$.}
    \label{tab:prederrors}
\end{table*}

\subsection{Predicting deep CNN classification accuracy using our Interpretable Models} \label{sec:pred_results}
Training datasets are often noisy; for example,  \cite{AN20} considers noisy datasets for hand-written character recognition.
We now exploit the proposed  interpretable model to predict how the deep CNN will perform with a  noisy training dataset without actually implementing the deep CNN.

Our predictive procedure is as follows.
We first train the CNNs on noisy datasets that are generated by adding simulated Gaussian noise with  noise variances chosen from a finite set. \footnote{Injecting artificial noise in training datasets is also used in variational autoencoders for robust feature learning~\cite{BNG09,VCN08}.} Then given the CNN decisions, we compute our interpretable model over this finite set of noise variances. Finally,  to predict how the  CNN will perform for a noise variance not in the set, we {\em interpolate} the utility function of the interpretable model at this noise variance. Then given the interpolated utility function and  information acquisition cost from our interpretable model, the predicted classification performance is computed by solving  convex optimization problem~\eqref{eqn:attentionmaximization}.
 The above procedure is formalized in Algorithm~\ref{alg:pred_interpolate}.
Hence, our interpetable model serves as a  computationally efficient method for predicting the performance of a CNN without implementing the CNN. The interpretable model can be viewed as a low dimension projection of the high-dimension  CNN with  predictive accuracy exceeding~94\%.

{\em Remark.} An  alternative procedure is  to directly interpolate the performance over the space of CNN weights (several hundreds of thousands). Due to the high dimensionality, this is an intractable interpolation. In comparison, interpolation over the utility functions in our interpretable model is over a few hundred variables.

\subsubsection*{Prediction Results of Algorithm~\ref{alg:pred_interpolate} on Deep CNN Performance} 
Table~\ref{tab:prederrors} displays the prediction errors (difference between the true and predicted classification accuracy) for the deep CNNs for all $5$ architectures and all image classes in CIFAR-10. For a fixed  CNN architecture and noise variance $\eta>0$, the prediction error $\delta_\eta(\state)$ for image class $\state$ is defined as:
\begin{equation}\label{eqn:pred_error_def}
    \delta_\eta(\state) =  |\hat{p}(\state|\state) - p_{\CNN}(\state|\state)|.
\end{equation}
In~\eqref{eqn:pred_error_def}, $\hat{p}(\cdot|\cdot)$ is the predicted CNN performance generated from Algorithm~\ref{alg:pred_interpolate} and $p_{\CNN}(\cdot|\cdot)$ is the true CNN performance obtained by implementing the CNN. Recall that $p(\state|\state)$ is the probability that the CNN correctly classifies an image belonging to class $\state$.

\begin{algorithm}[h]
\begin{algorithmic}
\REQUIRE Dataset $\datasetaccum$~\eqref{eqn:def_dataset} from $\numdp$ deep CNNs from a fixed network architecture. The  $\dpiter\uth$ CNN is trained on a noisy dataset with added Gaussian noise with variance $\eta_\dpiter=1+0.1\times(k-1)$.\\
\hspace{-0.35cm}\textbf{Step 1:} {\em Constructing Interpretable Model.}
The most robust  utility functions $\{\utilitysymbol_\dpiter^\ast\}_{\dpiter=1}^\numdp$ and information acquisition cost $\RIcost^\ast$ are computed by solving the following convex optimization problem.
\begin{align}
    &\hspace{-0.2cm}\{\utilitysymbol_\dpiter^\ast,\runcostinst^\ast_\dpiter\}_{\dpiter=1}^\numdp = \argmax_{\utilitysymbol_{1:\numdp}}\frac{\eps~\numdp}{\sum_{\dpiter=1}^{\numdp}\|\utilitysymbolagent{\dpiter}\|_2^2},~\text{\dtest}(\datasetaccum,\cdot) \leq -\eps.\nonumber\\
   & \hspace{-0.2cm}\RIcost^\ast(p(\action|\state))  = \max_{\dpiter=1} \runcostinst_\dpiter^\ast + \sum_{\state,\action}\prior(\state)(p(\action|\state)-p_\dpiter(\action|\state))\utilitysymbol^\ast(\state,\action)\nonumber.
\end{align}

\hspace{-0.35cm}\textbf{Step 2:} {\em Predicting Classification Accuracy.} For an arbitrary noise variance $\eta_1\leq \eta \leq \eta_\numdp$, obtain index $g,~1\leq g<\numdp$ such that $\eta\in[\eta_g,\eta_{g+1}]$. Then, the predicted classification accuracy $\hat{p}(\action|\state)$ for noise variance $\eta$ is computed as follows:
\begin{align}
    &\hspace{-0.1cm}\hat{p}(\action|\state)  = \argmax_{p(\action|\state)} \sum_{\action}\max_{\actiontwo}\sum_{\state}\prior(\state)p(\action|\state)\hat{\utilitysymbol}(\state,\action)-\RIcost^\ast(p),\nonumber\\
    &\hspace{-0.1cm}\hat{\utilitysymbol}  = 10\times  \{(\eta_{g+1}-\eta)\utilitysymbol^\ast_g + (\eta-\eta_g)\utilitysymbol^\ast_{g+1}\}.\label{eqn:pred_interpolate}
\end{align}
\hspace{-0.35cm}\textbf{Return:} Predicted CNN performance $\hat{p}(\action|\state)$ for noise variance $\eta$.
\end{algorithmic}
\caption{Predicting Deep CNN Classification Accuracy via the \RImodels\ model using Theorem~\ref{thrm:sparse}.}
\label{alg:pred_interpolate}
\end{algorithm}

\subsubsection*{Discussion and Insights}
(i) Our interpretable model can predict CNN classification performance with accuracy exceeding 94\% (see below).\\
(ii) The interpretable model (utility functions and information acquisition cost) for our predictive procedure (Algorithm~\ref{alg:pred_interpolate}) is evaluated on the set of noise variances $G_1=\{1+0.1\times(\dpiter-1),\dpiter=1,2,\ldots,11\}$. The predictive procedure of Algorithm~\ref{alg:pred_interpolate} is applied on the the set of noise variances given by $G_2=\{1.05+0.1\times(\dpiter-1),\dpiter=1,2,\ldots,10\}$. Table~\ref{tab:prederrors} displays the prediction errors $\delta_{\eta}(\state)$ averaged over all $\eta\in G_2$.
\\
(iii) From Table~\ref{tab:prederrors}, the prediction error $\delta_\eta(\state)$ averaged over all image classes $\state$ for the $5$ CNN architectures are:
\begin{multicols}{2}
\begin{compactenum}
    \item LeNet - $0.038$
    \item AlexNet - $0.036$
    \item VGG16 - $0.041$ 
    \item ResNet-50 - $0.027$ 
    \item NiN- $0.035$
\end{compactenum}
\end{multicols}
\noindent So the least accuracy is 95.9\%, and highest accuracy is 97.3\%. 
\\
(iv) The prediction error averaged over the network architectures was observed to be minimum
for image class `cat' (98.1\%)  and maximum for image class `deer' (95.7\%) over all image classes.\\
(iv) {\em Statistical Similarity between Deep CNNs and Interpretable Model.} We computed the Kullback-Leibler (KL) divergence between the true and predicted classification performances $p_{\text{imp}}(\action|\state)$ and $\hat{p}(\action|\state)$. Recall $\hat{p}(\action|\state)$ is computed from the interpretable model via Algorithm~\ref{alg:pred_interpolate} and $p_{\CNN}(\action|\state)$ is obtained from the CNN.
The KL divergence values for the 5 CNN architectures are:
\begin{multicols}{2}
\begin{compactenum}
\item LeNet - $0.015$ \item AlexNet - $0.012$ \item VGG16 - $0.016$ \item Resnet-50 - $0.006$ \item NiN - $0.018$.
\end{compactenum}
\end{multicols}
\noindent Thus the decisions made by the deep CNNs are statistically similar  to decisions generated by our interpretable model.

\section{Conclusions and Extensions}
This paper has developed an interpretable model for deep image classification using micro- and behavioral economics theory. By extensive analysis of the decisions of $200$ CNNs over $5$ popular CNN architectures, we showed that deep CNNs can be explained by rationally inattentive Bayesian utility maximization.

Our main results were the following: 
\\
{\em 1.} Using the theory of Bayesian revealed preference,  Theorem~\ref{thrm:IRL_DNN} gave a necessary and sufficient condition for the actions of a collection of decision makers to be consistent with rationally inattentive  Bayesian utility maximization. We showed that deep CNNs operating on the CIFAR-10 dataset satisfy these necessary and sufficient conditions.\\
{\em 2.} Next  we studied the robustness margin by which the deep CNNs satisfy Theorem~\ref{thrm:IRL_DNN}; we found that the margins were sufficiently large implying robustness of the results. Our robustness results are summarized in Table~\ref{tab:res_rob_dtest_both}.  \\
{\em 3.} In Theorem~\ref{thrm:sparse}, we constructed the sparsest interpretable model from the feasible set generated using Theorem~\ref{thrm:IRL_DNN}. The sparsest interpretable model explains deep CNN decisions using the least number of parameters. The sparsest interpretable model introduces a useful preference ordering amongst
the set of hypotheses (image labels) considered by the deep
neural network; for example, how much additional priority
is allocated to the classification of a cat as a cat compared
to a cat as a dog. In classical deep learning, this preference
ordering is not explicitly generated
\\
{\em 4.} Finally, we showed that our interpretable model can predict CNN performance with accuracy exceeding 94\%, and the decisions generated by our interpretable model are statistically similar to that of a deep CNN. 
At a more conceptual level, our results suggest that  deep CNNs for image classification  are equivalent to an economics-based constrained Bayesian decision system (used in micro-economics  to model human decision making).

{\em Extensions.} An immediate extension of this work is to construct appropriately designed image features to replace the image class label as the state. This would result in a richer descriptive model of the  CNN due to more degrees of freedom in the utility function.

Our proposed interpretable model generates a concave  utility function by design. This is an important feature of the revealed preference framework; even though the actual deep learner's utility may not be convex.
To quote Varian~\cite{VR82}: "If data can be rationalized by any non-trivial utility function, then it can be rationalized by a nice utility function. Violations of concavity  cannot be detected with only a finite number of observations." 
 A more speculative  extension is to investigate the asymptotic behavior of the \dtest\ and \dtestcompact\ decision tests for rationally inattentive utility maximization-do the tests pass when the number of deep CNNs tend to infinity? Recent results \cite{REN15} show that an infinite dataset can at best be rationalized  by a quasi-concave utility function.

{\bf Reproducibility}: The computer programs and deep image classification datasets needed to reproduce all the results in this paper can be obtained from the public GitHub repository \url{https://github.com/KunalP117/DL_RI}. 

\section*{Acknowledgement}
This research was supported in part by the Army Research Office under grants W911NF-21-1-0093 and W911NF-19-1-0365, and the National Science Foundation under grant CCF-2112457.
\bibliographystyle{unsrt_abbrv_custom}
\bibliography{BRPCNN}


\section*{Appendix}

\renewcommand{\thesection}{A}

\subsection{Proof of Theorem~\ref{thrm:IRL_DNN}}\label{proof:Thrm_BRP}
{\it Proof of necessity of NIAS and NIAC:}	
 \begin{compactenum}
 	\item NIAS~(\ref{eqn:NIAS}): For agent $\dpiter\in\dpset$, define the subset $\obsset_{\action}\subseteq\obsset$ so that for any observation $\obs\in\obsset_{\action}$, given posterior pmf $p_{\dpiter}(\state|\obs)$, the optimal choice of action is $\action$ \eqref{eqn:utilitymaximization}. We define the revealed posterior pmf given action $\action$ as $p_{\dpiter}(\state|\action)$. The revealed posterior pmf is a stochastically garbled version of the actual posterior pmf $p_{\dpiter}(\state|\obs)$, that is,
 	\begin{equation}\label{eqn:revpos}
 	  p_{\dpiter}(\state|\action) = \sum_{\obs\in\obsset} \frac{p_{\dpiter}(\state,\obs,\action)}{p_{\dpiter}(\action)} = \sum_{\obs\in\obsset} p_{\dpiter}(\obs|\action)p_{\dpiter}(\state|\obs)
 	\end{equation} 
 	Since the optimal action is $a$ for all $\obs\in\obsset_{\action}$, \eqref{eqn:utilitymaximization} implies:
 	\begin{align*}
 	& \hspace{-0.4cm}\quad\quad~~\sum_{\state\in \stateset} p_{\dpiter}(\state|\obs) (\utilitysymbolagent{\dpiter}(\state,\actiontwo)-\utilityagent{\dpiter})\leq 0\\
 	&\hspace{-0.4cm}\implies \sum_{\obs\in \obsset_{\action} } p_{\dpiter}(\obs|\action) \sum_{\state\in \stateset} p_{\dpiter}(\state|\obs) (\utilitysymbolagent{\dpiter}(\state,\actiontwo)-\utilityagent{\dpiter})\leq 0 \\
    & \hspace{-0.4cm}\implies \sum_{\obs\in \obsset} p_{\dpiter}(\obs|\action) \sum_{\state\in \stateset} p_{\dpiter}(\state|\obs) (\utilitysymbolagent{\dpiter}(\state,\actiontwo)-\utilityagent{\dpiter})\leq 0 \\
    &\hspace{-0.4cm}\quad\quad~~(\text{since }p_{\dpiter}(\obs|\action)=0,~\forall \obs\in\obsset\backslash\obsset_{\action})\\
  	&\hspace{-0.4cm}\implies \sum_{\state\in\stateset} \sum_{\obs\in \obsset} p_{\dpiter}(\obs|\action) p_{\dpiter}(\state|\obs) (\utilitysymbolagent{\dpiter}(\state,\actiontwo)-\utilityagent{\dpiter})\leq 0 \\
	&\hspace{-0.4cm}\implies\sum_{\state\in \stateset} p_{\dpiter}(\state|\action) (\utilitysymbolagent{\dpiter}(\state,\actiontwo)-\utilityagent{\dpiter})\leq 0~(\text{from }\eqref{eqn:revpos})
 	\end{align*}  
 	This is precisely the NIAS inequality~\eqref{eqn:NIAS}. 
 	\item NIAC~(\ref{eqn:NIAC}): Let $\runcostinst_{\dpiter}=\RIcost(\attfunsymb_{\dpiter})>0$, where $\RIcost(\cdot)$ denotes the information acquisition cost of the collection of agents $\dpset$. Also, let $J(\attfunagent{\dpiter},\utilitysymbolagent{\dpiter})$ denote the expected utility of the $\dpiter\uth$ agent given attention strategy $\attfunagent{\dpiter}$ (first term in RHS of \eqref{eqn:attentionmaximization}). Here, the expectation is taken wrt both the state $\state$ and observation $\obs$. It can be verified that $J(\cdot,\utilitysymbol_\dpiter)$ is convex in the first argument. Finally, for the $\dpiter\uth$ agent, we define the revealed attention strategy $\attfunagent{\dpiter}'$ over the set of actions $\actionset$ as
 	\begin{equation*}
 	    \attfunagent{\dpiter}'(\action|\state) = p_{\dpiter}(\action|\state),~\forall\action\in\actionset,
 	\end{equation*}
 	where the variable $p_{\dpiter}(\action|\state)$ is obtained from the dataset $\datasetaccum$. Clearly, the revealed attention strategy is a stochastically garbled version of the true attention strategy since
 	\begin{equation}\label{eqn:revattfun}
 	    \attfunagent{\dpiter}'(\action|\state) = p_{\dpiter}(\action|\state) = \sum_{\obs\in\obsset}p_{\dpiter}(\action|\obs)\attfunagent{\dpiter}(\obs|\state)
 	\end{equation}
 	From Blackwell dominance~\cite{BW53} and the convexity of the expected utility functional $J(\cdot,\utilitysymbol_\dpiter)$,  it follows that: 
 	\begin{equation}
 	    J(\attfunsymb_\dpiter',\utilitysymbol_\dpitertwo) \leq J(\attfunsymb_\dpiter,\utilitysymbol_\dpitertwo), \label{eqn:ineq_BLK}
 	\end{equation}
 	when $\attfunsymb_\dpiter$ Blackwell dominates $\attfunsymb_\dpiter'$. The above relationship holds with equality if $\dpiter=\dpitertwo$ (this is due to NIAS \eqref{eqn:NIAS}).
We now turn to condition \eqref{eqn:attentionmaximization} for optimality of attention strategy.
The following inequalities hold for any pair of agents $\dpitertwo\neq\dpiter$:
    \begin{align}
        &J(\attfunsymb_\dpiter',\utilitysymbolagent{\dpiter}) - \runcostinst_{\dpiter}\overset{\eqref{eqn:ineq_BLK}}{=} J(\attfunagent{\dpiter},\utilitysymbolagent{\dpiter}) - \runcostinst_{\dpiter}\nonumber\\
        \overset{\eqref{eqn:attentionmaximization}}{\geq}~&J(\attfunagent{\dpitertwo},\utilitysymbolagent{\dpiter}) - \runcostinst_{\dpitertwo} \overset{\eqref{eqn:ineq_BLK}}{\geq}~J(\attfunagent{\dpitertwo}',\utilitysymbolagent{\dpiter}) - \runcostinst_{\dpitertwo}\label{eqn:NIAC_proof}.
    \end{align}
    This is precisely the NIAC inequality~\eqref{eqn:NIAC}.
\end{compactenum}

\noindent{\em Proof for sufficiency of NIAS and NIAC:}
Let $\{\utilitysymbol_\dpiter,\runcostinst_\dpiter\}_{\dpiter=1}^\numdp$ denote a feasible solution to the NIAS and NIAC inequalities of Theorem~\ref{thrm:IRL_DNN}. To prove sufficiency, we construct an \RImodel\ tuple as a function of dataset $\datasetaccum$ and the feasible solution that satisfies the optimality conditions~\eqref{eqn:utilitymaximization},\eqref{eqn:attentionmaximization} of Definition~\ref{def:RI}.
 
Consider the following \RImodel\ model tuple:
 \begin{align}
     &\bigtuple  = (\dpset,\stateset,\obsset=\actionset,\actionset,\prior,\RIcost,\{p_\dpiter(\action|\state),\utilitysymbol_\dpiter,\dpiter\in\dpset\}),\text{ where}\nonumber\\
     &\RIcost(p(\action|\state))  = \max_{\dpiter\in\dpset} \runcostinst_\dpiter + J(p(\action|\state),\utilitysymbol_\dpiter) - J(p_\dpiter(\action|\state),\utilitysymbol_\dpiter).\label{eqn:construct_cost}
 \end{align}
 In \eqref{eqn:construct_cost}, $C(\cdot)$ is a convex cost since it is a point-wise maximum of monotone convex functions. Further, since NIAC is satisfied, \eqref{eqn:construct_cost} implies $\RIcost(\attfunagent{\dpiter})=\runcostinst_{\dpiter}$. It only remains to show that inequalities \eqref{eqn:utilitymaximization} and \eqref{eqn:attentionmaximization} in Definition~\ref{def:RI} are satisfied for all agents in $\dpset$.
 \begin{compactenum}
 \item {\em NIAS implies \eqref{eqn:utilitymaximization} holds.} This is straightforward to show since the observation and action sets are identical.
 \item {\em Information Acquisition Cost \eqref{eqn:construct_cost} implies \eqref{eqn:attentionmaximization} holds.} Fix agent $\dpitertwo\in\dpset$. Then, for any attention strategy $p(\action|\state)$, the following inequalities hold.
 \begin{align*}
     &\RIcost(p(\action|\state))  =  \max_{\dpiter\in\dpset}~ \runcostinst_{\dpiter} + J(p(\action|\state),\utilitysymbolagent{\dpiter}) - J(p_\dpiter(\action|\state),\utilitysymbolagent{\dpiter}) \\
     &\hspace{-0.4cm}\implies J(p_\dpitertwo(\action|\state))  -\runcostinst_\dpitertwo \geq J(p(\action|\state)) - C(p(\action|\state)),~\forall~p(\action|\state)\\
     &\hspace{-0.4cm}\implies p_\dpiter(\action|\state)\in \argmax_{p(\action|\state)} J(p(\action|\state),\utilitysymbolagent{\dpiter}) - \RIcost(p(\action|\state))=\eqref{eqn:attentionmaximization}.
 \end{align*}
 \end{compactenum}

\subsection{\RImodels\,(Sparse \RImodel\,) Model for Rationally Inattentive Bayesian Utility Maximization}\label{sec:umri_model_compact}
In Sec.\,\ref{sec:umri_model}, we outlined the \RImodel\ model for rationally inattentive utility maximization of $\numdp$ Bayesian agents parameterized by $\numdp$ utility functions and a cost of information acquisition. This section proposes a sparse version of the \RImodel\ model, namely, the \RImodels\ model that is parameterized by a {\em single} utility function that rationalizes the decisions of $\numdp$ Bayesian agents.
Abstractly, the \RImodels\ model is described by the tuple
\begin{equation}
\bigtuple = (\dpset,\stateset,\obsset,\actionset,\prior,\RIcost,\utilitysymbol,\{\attfunagent{\dpiter},\lambda_\dpiter,\dpiter\in\dpset\}).
    \label{eqn:RI_tuple_compact}
\end{equation}
All parameters in \eqref{eqn:RI_tuple_compact} are identical to that in \eqref{eqn:RI_tuple} except for the additional parameter $\lambda_\dpiter
\in\reals_+$. $\lambda_\dpiter$ can be interpreted as the sensitivity to information acquisition of the $\dpiter\uth$ agent. We discuss the significance of $\lambda_\dpiter$ in more detail below.
In complete analogy to Definition~\ref{def:RI}, Definition~\ref{def:RI_compact} below specifies the optimal action and attention strategy policy of the Bayesian agents in $\dpset$.

\begin{definition}[Rationally Inattentive Utility Maximization for \RImodels\,]
\label{def:RI_compact} Consider a collection of Bayesian agents $\dpset$ parameterized by $\bigtuple$ in \eqref{eqn:RI_tuple_compact} under the \RImodels\ model. Then,\\
\noindent (a) {\bf Expected Utility Maximization:}  
Given posterior pmf $p(\state|\obs)$, agent $\dpiter\in\dpset$ chooses action $\action$ that maximizes its  expected utility.
\begin{equation}
\action \in\argmax_{\action' \in \actionset}~\E_{\state}\{ \utilitysymbol_\dpiter(\state,\action') | \obs\}= \sum\limits_{\state\in\stateset}p(\state|\obs)\utilitysymbol(\state,\action')
\label{eqn:utilitymaximization_compact}
\end{equation}
\noindent (b) {\bf Attention Strategy Rationality:} Agent $\dpiter$ chooses attention strategy $\attfunagent{\dpiter}$ that optimally trades off between utility maximization and cost minimization.
\begin{align}
\attfunsymb_\dpiter \in\argmax_{\attfunsymb'} 
\E_{\obs}\{&\operatorname*{max}_{\action\in\actionset}\E_{\state}\{ \utilitysymbol(\state,\action) | \obs\}\}- \lambda_\dpiter\RIcost(\attfunsymb',\prior)
\label{eqn:attentionmaximization_compact}
\end{align}
\end{definition}
{\em Remarks.} 1. {\em Role of $\lambda_\dpiter$.} In \eqref{eqn:attentionmaximization_compact}, $\lambda_\dpiter$ is the differentiating parameter across agents. Even though all agents have the same utility function, different values of $\lambda_{\dpiter}$ result in different optimal strategies $\attfunagent{\dpiter}$~\eqref{eqn:attentionmaximization_compact}. \\
2. {\em Sparsity of \RImodels\,.} The \RImodel\ model tuple for $\numdp$ Bayesian agents is parameterized using $\numdp(|\stateset||\actionset| + 1)$ variables. In comparison, the \RImodels\ tuple is parameterized via $|\stateset||\actionset|+\numdp$ variables. The difference in variables for parametrization is linear in $\numdp$.


Finally, in complete analogy to Theorem~\ref{thrm:IRL_DNN}, we now state Theorem~\ref{thrm:IRL_DNN_compact} that states necessary and sufficient conditions for the decisions of a collection of Bayesian agents to be rationalized by the \RImodels\ model.
\begin{theorem}[\dtestcompact\ Test for Rationally Inattentive Utility Maximization]\label{thrm:IRL_DNN_compact} Given the dataset $\datasetaccum$ \eqref{eqn:dataset_accum} obtained from a collection of Bayesian agents $\dpset$. Then,\\
{\em 1.} \underline{Existence:}  There exists a \RImodels\ tuple $\bigtuple(\datasetaccum)$ \eqref{eqn:RI_tuple} that rationalizes dataset $\datasetaccum$ if and only if there exists a feasible solution that satisfies the set of inequalities 
\begin{equation}\label{eqn:BRP_ineq_compact}
\text{\dtestcompact}(\datasetaccum) \leq \mathbf{0}.
\end{equation}
In \eqref{eqn:BRP_ineq}, \dtestcompact$(\cdot)$ corresponds to a set of inequalities stated in Algorithm~\ref{alg:dtest_compact} below. The set-valued estimate of $\bigtuple$ that rationalizes $\datasetaccum$ is the set of all feasible solutions to \eqref{eqn:BRP_ineq}.\\
{\em 2.} \underline{Reconstruction:} Given a feasible solution $\{\utilitysymbol,\lambda_\dpiter,\runcostinst_\dpiter\}_{\dpiter=1}^\numdp$ to $\dtestcompact(\datasetaccum,\cdot)$, $\utilitysymbol$ is the $\dpiter\uth$ Bayesian agent's utility function, for all $\dpiter=1,2,\ldots,\numdp$. The set of observations $\obsset=\actionset$, the set of actions in $\datasetaccum$. The feasible cost of information acquisition $\RIcost$ in $\bigtuple(\datasetaccum)$ is defined in terms of the feasible variables $\runcostinst_\dpiter,\lambda_\dpiter$ as: 
\begin{align}
    \RIcost(\attfunsymb) & = \max_{\dpiter\in\dpset} \runcostinst_\dpiter + \lambda_\dpiter \sum_{\state,\action}(p(\state,\action)-p_{\dpiter}(\state,\action))\utilitysymbol(\state,\action),\nonumber\\
    &~(\attfunsymb=\{p(\action|\state),\action\in\actionset,\state\in\stateset\})\label{eqn:reconstruct_cost_compact}
\end{align}
\end{theorem}
\noindent The proof of Theorem~\ref{thrm:IRL_DNN_compact} closely follows the proof of Theorem~\ref{thrm:IRL_DNN} and hence, omitted. In comparison to the \dtest\ test of Theorem~\ref{thrm:IRL_DNN}, the \dtestcompact\ test has the same number of inequalities but fewer decision variables. 
Hence, the set of feasible parameters generated from Algorithm~\ref{alg:dtest_compact} is smaller compared to Algorithm~\ref{alg:dtest}.
\begin{algorithm}
\begin{algorithmic}
\REQUIRE Dataset $\datasetaccum=\{\prior,p_{\dpiter}(\action|\state),\state,\action\in\stateset,\dpiter\in\dpset\}$ from a collection of Bayesian agents $\dpset$.

\hspace{-0.35cm}\textbf{Find:} Positive reals $\runcostinst_{\dpiter},\lambda_\dpiter,\utilitysymbol\in(0,1]$ for all $\state\in\stateset,$ $\action\in\actionset,~\dpiter\in\dpset$ that satisfy the following inequalities:
\begin{align*}
    &\hspace{-0.8cm}1.~\sum_{\state}p_{\dpiter}(\state|\action)~(\utilitysymbol(\state,\actiontwo) -\utilitysymbol(\state,\action))\leq 0, \forall\action,\actiontwo,\dpiter,\\ 
    &\hspace{-0.8cm}2.~\sum_{\state,\action} (p_{\dpitertwo}(\state,\action)-p_\dpiter(\state,\action))\utilitysymbol(\state,\action)+\lambda_\dpiter(\runcostinst_{\dpiter}-\runcostinst_{\dpitertwo})\leq 0,\forall\dpitertwo,\dpiter,
\end{align*}
\hspace{-0.14cm}where $p_{\dpiter}(\state,\action)=\prior(\state)p_{\dpiter}(\action|\state),~p_{\dpiter}(\state|\action)=\frac{p_{\dpiter}(\state,\action)}{\sum_{\state'}p_{\dpiter}(\state',\action)}$.\\\vspace{0.1cm}

\hspace{-0.35cm}\textbf{Return:} Set of feasible utility function $\utilitysymbol$, scalars $\lambda_\dpiter$ and information acquisition costs $\runcostinst_{\dpiter}$ incurred by agents $\dpiter\in\dpset$.
\end{algorithmic}
\caption{\dtestcompact\ Convex Feasibility Test of Theorem~\ref{thrm:IRL_DNN_compact}}
\label{alg:dtest_compact}
\end{algorithm}

\subsection{Construction of Deep CNN Dataset}
\label{sec:construct_dataset}
We now explain how the decisions of the deep CNNs are incorporated into our main theorems
Theorems~\ref{thrm:IRL_DNN} and \ref{thrm:IRL_DNN_compact}.
%
Suppose $\numdp$ deep CNNs indexed by $\dpiter=1,\ldots,\numdp$ with different training parameters are trained on the CIFAR-10 dataset. For every trained deep CNN $\dpiter$, given test image $\imgiter$ from CIFAR-10 test dataset with image class $\testimg_\imgiter$, let the vector $\testpred_{\imgiter,\dpiter}\in\Delta^{9}$ denote the corresponding softmax output of the deep CNN. The vector $\testpred_{\imgiter,\dpiter}$ is a $10$-dimensional probability vector where $\testpred_{\imgiter,\dpiter}(j)$ is the probability that deep CNN $\dpiter$ classifies test image $\imgiter$ into image class $j$.

The decisions of all $\numdp$ deep CNNs on the CIFAR-10 test dataset are aggregated into dataset $\datasetaccum$ for compatibility with Theorems~\ref{thrm:IRL_DNN} and \ref{thrm:IRL_DNN_compact} as follows:
\begin{align}
    &\datasetaccum=\{\prior,p_{\dpiter}(\action|\state),\state,\action\in\stateset,\dpiter\in\{1,2,\ldots,\numdp\}\},\text{ where}\nonumber\\
     &\prior(\state) = \sum_{\imgiter=1}^{\numimg} \frac{\mathbbm{1}\{\testimg_{\imgiter}=\state\}}{\numimg},~p_{\dpiter}(\action|\state)= \frac{\sum_{\imgiter=1}^{\numimg} \mathbbm{1}\{\testimg_{\imgiter}=\state\}\testpred_{\imgiter,\dpiter}(\action) }{\sum_{\imgiter=1}^{\numimg}\mathbbm{1}\{ \testimg_{\imgiter}=\state\}},\nonumber\\
    & \numimg=10000,~\stateset = \actionset = \{1,2,\ldots 10\}.\label{eqn:def_dataset}
\end{align}
Here $\prior(\state)$ is the empirical probability that the image class of a test image in the CIFAR-10 test dataset is $\state$. Since the output of the CNN is a probability vector, we compute $p_{\dpiter}(\action|\state)$ for the $\dpiter\uth$ CNN by averaging the $\action\uth$ component of the output over all test images in image class $\state$. Finally, $\numimg$ is the number of test images in the CIFAR-10 test dataset, and the set of true and predicted image classes are the same, i.e., $\stateset=\actionset$. Although implicit in the above description, our Bayesian revealed preference approach to interpretable deep image classification assumes the deep CNN's (agent's) ground truth is the true image label, and its decision $\action$ is the predicted image label.

\end{document}